\documentclass[10pt,twocolumn,letterpaper]{article}

\usepackage[pagenumbers]{cvpr} 

\usepackage{lipsum}
\usepackage{mwe}
\usepackage{graphicx}
\usepackage{multirow}
\usepackage{url}                
\usepackage{amsfonts}
\usepackage{nicefrac}
\usepackage{microtype}  
\usepackage{xcolor}
\usepackage{amsmath}
\usepackage{makecell}

\usepackage[table]{xcolor}
\usepackage{caption,subcaption}

\definecolor{gan}{RGB}{0,112,192}
\definecolor{diff}{RGB}{78,167,46}
\definecolor{ours}{RGB}{192,0,0}

\definecolor{cvprblue}{rgb}{0.21,0.49,0.74}
\usepackage[pagebackref,breaklinks,colorlinks,allcolors=cvprblue]{hyperref}

\title{Revisiting the Perception-Distortion Trade-off with Spatial-Semantic Guided Super-Resolution} 

\author{
Dan Wang$^{1*}$ \quad Haiyan Sun$^{2*}$ \quad Shan Du$^{3}$ \quad Z. Jane Wang$^{3}$ \\
Zhaochong An$^{4}$ \quad Serge Belongie$^{4}$ \quad Xinrui Cui$^{2}$ \\
$^{1}$University of California, San Diego \quad
$^{2}$University of North Texas \\
$^{3}$University of British Columbia \quad
$^{4}$University of Copenhagen
}

\begin{document}

\twocolumn[{
\maketitle
\begin{center}
    \includegraphics[width=0.9\linewidth]{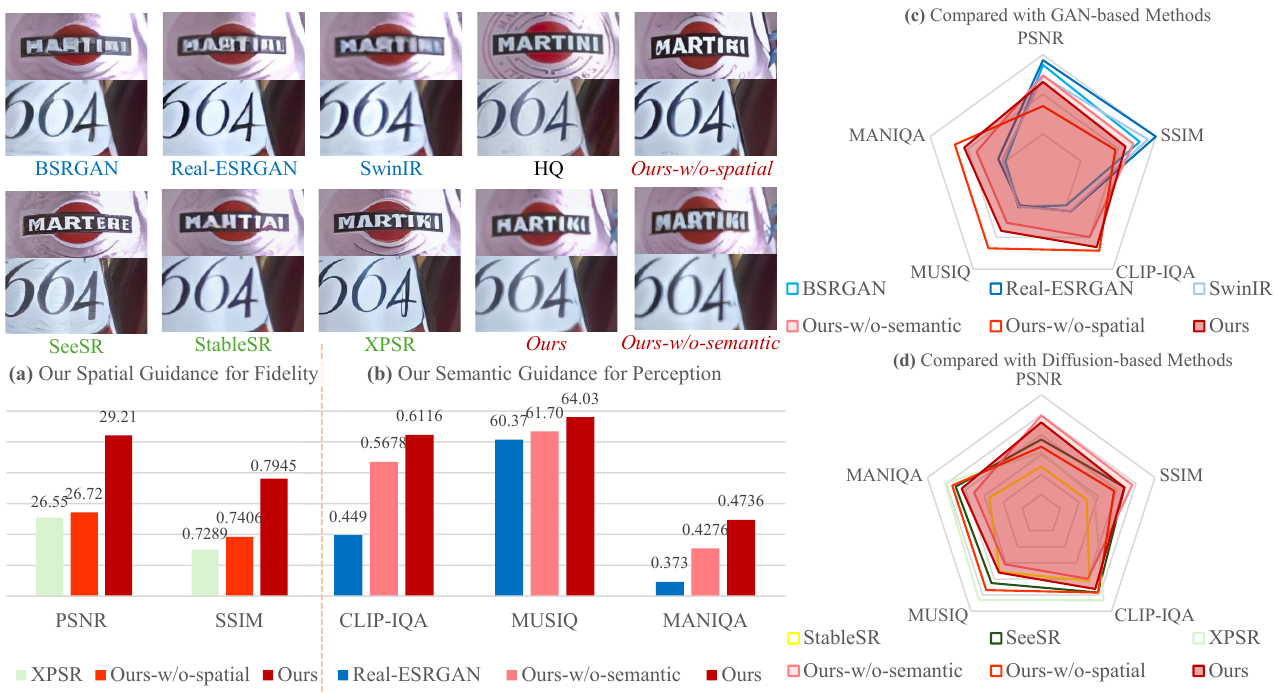}
\end{center}
\vspace{-10pt}
\captionof{figure}{Perception-distortion trade-off in \textcolor{gan}{GAN-based}, \textcolor{diff}{Diffusion-based}, and \textcolor{ours}{\textit{Ours}}: GAN-based methods reduce distortion but produce blurry textures, while diffusion-based methods generate perceptually sharp yet hallucinated details. By integrating spatial-grounded textual guidance, SpaSemSR improves reconstruction fidelity (PSNR, SSIM in (a)), while semantic-enhanced visual guidance enhances perceptual quality (CLIP-IQA, MUSIQ, MANIQA in (b)), resulting in a better perception-distortion trade-off compared with GAN-based (c) and diffusion-based models (d).}
\label{fig:teaser}
\vspace{10pt}
}]

{
\renewcommand{\thefootnote}{\fnsymbol{footnote}}
\footnotetext[1]{Equal contribution.}
}

\begin{abstract}
Image super-resolution (SR) aims to reconstruct high resolution images with both high perceptual quality and low distortion, but is fundamentally limited by the perception–distortion trade-off. GAN-based SR methods reduce distortion but still struggle with realistic fine-grained textures, whereas diffusion-based approaches synthesize rich details but often deviate from the input, hallucinating structures and degrading fidelity. This tension raises a key challenge: how to exploit the powerful generative priors of diffusion models without sacrificing fidelity. To address this, we propose \textbf{SpaSemSR}, a spatial–semantic guided diffusion framework with two complementary guidances. First, \textbf{spatial-grounded textual guidance} integrates object-level spatial cues with semantic prompts, aligning textual and visual structures to reduce distortion. Second, \textbf{semantic-enhanced visual guidance} with a multi-encoder design and semantic degradation constraints unifies multimodal semantic priors, improving perceptual realism under severe degradations. These complementary guidances are adaptively fused into the diffusion process via spatial-semantic attention, suppressing distortion and hallucination while retaining the strengths of diffusion models. Extensive experiments on multiple benchmarks show that SpaSemSR achieves a superior perception-distortion balance, producing both realistic and faithful restorations.
\url{https://hssmac.github.io/SpaSemSR_web/}
\end{abstract}
\section{Introduction}
\label{sec:intro}

Image Super-Resolution (SR) aims to reconstruct high-resolution (HR) images from low-resolution (LR) inputs degraded by complex and often unknown processes, with the dual objective of achieving high fidelity and strong perceptual quality. However, as demonstrated by ~\cite{blau2018perception}, SR methods are fundamentally constrained by the \textbf{Perception-Distortion trade-off}: improving distortion (i.e., reconstruction accuracy) inevitably comes at the cost of perceptual quality, and vice versa, due to the monotone boundary between the two. Distortion is measured by full-reference metrics such as PSNR and SSIM, while perception emphasizes visual realism regardless of ground-truth similarity, and is assessed by reference-free metrics such as CLIP-IQA, MUSIQ, and MANIQA.

GAN-based methods ~\cite{liang2022details, zhang2022perception, zhang2021designing, wang2021real} have demonstrated strong fidelity performance, but their gains in reconstruction accuracy do not always translate into perceptual improvements (e.g., lower CLIP-IQA, MUSIQ). As illustrated in Fig.~\ref{fig:teaser}, these models often produce artifacts or blurring issues, stemming from unstable adversarial training, domain shifts between synthetic training and real-world test data, and fidelity-biased optimization objectives. Recently, diffusion-based approaches ~\cite{wang2024exploiting, lin2024diffbir, yang2024pixel, wu2024seesr, qu2024xpsr, chen2025faithdiff} have emerged as powerful alternatives. Leveraging the rich generative priors of large-scale text-to-image (T2I) diffusion models, they excel in generating realistic textures and achieve superior scores on perceptual quality metrics. Yet, somewhat counter-intuitively, these methods often underperform GAN-based models on distortion metrics (e.g., lower PSNR and SSIM), leading to reduced fidelity and frequent hallucinations, as shown in Fig.~\ref{fig:teaser}. More recently, lots of studies~\cite{wu2024one, dong2025tsd, zhang2025time} have shifted their focus toward improving the efficiency of diffusion models, aiming to design more effective one-step generation schemes for Stable Diffusion, thereby reducing the heavy computational cost caused by the multi-step denoising process. However, these approaches do not primarily focus on addressing the distortion problem caused by hallucinatory artifacts generated in diffusion models. Thus, neither paradigm fully resolves the tension between perception and distortion, leaving it as an unsolved issue.

To further investigate the Perception-Distortion trade-off, we revisit the mechanisms underlying diffusion-based SR. Despite their strong generative priors, diffusion models face three key challenges: (1) Their generative nature promotes sample diversity, which benefits image synthesis but undermines fidelity in restoration. (2) Severe degradations in LR inputs often destroy local structures, leading to ambiguous semantics. (3) Existing solutions, such as PASD~\cite{yang2024pixel}, SeeSR~\cite{wu2024seesr}, address this by introducing semantic text prompts as auxiliary conditions. However, text prompts alone lack spatial awareness, and thus reconstructions remain distorted compared to the ground truth, yielding high perceptual but low fidelity scores ~\cite{ren2025hallucination}. 
 
In this work, we present \textbf{Spa}tial-\textbf{Sem}antic guided SR framework (SpaSemSR), a diffusion-based approach that pushes the perception-distortion boundary by integrating complementary spatial and semantic guidance. Specifically, we introduce: (1) \textbf{spatial-grounded textual guidance}, which aligns spatial information with semantic textual prompts to improve fidelity (Fig.~\ref{fig:teaser} (a)). (2) \textbf{semantic-enhanced visual guidance}, which enriches perceptual quality (Fig.~\ref{fig:teaser} (b)) by constraining generation with multimodal semantic priors. Together, two forms of guidance bridge the gap between semantically rich but spatially ambiguous text prompts and spatially precise but semantically degraded visual features.

To achieve it, \textbf{first}, we introduce a novel spatial-aware text fusion mechanism for better fidelity representation that integrates object-level spatial coordinates with corresponding semantic textual tags. \textbf{Second}, to extract rich and robust semantics from degraded LR inputs, we design a two-branch image encoder system: one captures low-level latent structures, while the other extracts high-level semantic features. The perception ability of these extracted features is further ensured by our novel semantic degradation constraints derived from pre-trained VAE~\cite{rombach2022high} and SAM~\cite{kirillov2023segment} encoders. \textbf{Third}, to effectively integrate semantic and spatial priors with diffusion generative priors, we propose Spatial-Semantic ControlNet, Spatial-aware Text Attention (SpaTextAtten), and Semantic-enhanced Image Attention (SemImgAtten) layers, enabling adaptive fusion of spatial and semantic guidance across modalities in the diffusion model. As a result, this spatial and semantic guidance yields reconstructions that are both perceptually compelling and faithful to the original content, thereby achieving a balance between perception and fidelity (Fig.~\ref{fig:teaser} (c) and (d)). 
Our contributions are summarized as: 
\begin{itemize}
\item[1.] We propose SpaSemSR, a novel diffusion-based framework for balancing perception and distortion, which jointly exploits complementary spatial and semantic guidance, curbing distortion and hallucination while fully leveraging the strengths of diffusion priors.
\item[2.] We introduce a spatial-aware text fusion mechanism that augments semantic prompts with spatial grounding, thereby improving generation fidelity and alleviating spatial misalignment between textual and visual representations.
\item[3.] We design a semantic-enhanced multi-encoder architecture with semantic degradation constraints that jointly capture low-level structures and high-level semantics, further constrained by pretrained VAE and SAM priors to provide robust perceptual learning. 
\item[4.] We propose the Spatial-Semantic ControlNet, SpaTextAtten, and SemImgAtten layers to effectively integrate semantic and spatial guidance into diffusion-based generation.
\item[5.] Extensive experiments demonstrate that SpaSemSR substantially reduces blurry textures and hallucinatory artifacts, achieving a state-of-the-art balance between perception and distortion across multiple benchmarks. Ablation studies further validate the complementary contributions of spatial and semantic guidance to fidelity and perceptual quality.
\end{itemize}
\section{Related Work}
\label{sec:related}

\paragraph{Image Super-Resolution.}
Classical SR methods~\cite{gu2019blind, huang2020unfolding, zhang2018learning} estimate predefined degradation kernels to recover high-resolution images. While effective on synthetic data, they struggle with complex real-world degradations. To address this, BSRGAN~\cite{zhang2021designing} introduced randomized degradation pipelines, and Real-ESRGAN~\cite{wang2021real} proposed high-order degradations with Sinc filters. GAN-based SR further incorporated perceptual losses to enhance visual quality, but often introduces artifacts and fails to reconstruct faithful textures, motivating the use of stronger generative priors.
\paragraph{Diffusion Prior-based Super-Resolution.}
Diffusion models leverage powerful generative priors to produce perceptually realistic SR results. StableSR~\cite{wang2024exploiting} fine-tuned Stable Diffusion with a Time-aware Encoder and controllable feature wrapping. DiffBIR~\cite{lin2024diffbir} used restoration modules and IRControlNet to remove degradations while preserving fidelity. Text-guided SR methods PASD~\cite{yang2024pixel}, SeeSR~\cite{wu2024seesr}, FaithDiff~\cite{chen2025faithdiff}, XPSR~\cite{qu2024xpsr} extract semantic cues from images or multimodal models to guide generation. However, these approaches often use global or loosely aligned text prompts, lacking precise spatial grounding. Our work explicitly integrates spatial-aware semantic guidance into diffusion for fidelity restoration.
\paragraph{Perception-Distortion Trade-off.}
Balancing perceptual realism and fidelity is a key SR challenge.
The perception-distortion trade-off was formalized in~\cite{blau2018perception}. Subsequent works addressed it via multi-objective strategies: a two-stage fidelity-then-perception pipeline ~\cite{zhang2022perception}, Bayesian optimization for dynamic loss weighting ~\cite{zhu2024perceptual}, and multi-objective optimization strategies ~\cite{sun2024perception}.
GAN-based methods typically favor fidelity but produce over-smoothed outputs, whereas diffusion-based models generate sharp textures at the risk of hallucinations. Existing balancing techniques are mostly GAN-centric. Our approach bridges this gap by constraining T2I diffusion with spatially grounded semantic guidance, improving fidelity while preserving perceptual quality.
\section{Methodology}
\label{sec:method}
\subsection{Motivation and Framework Overview}
\label{sec:framework}

\begin{figure*}[t]
    \centering
    \includegraphics[width=0.9\textwidth]{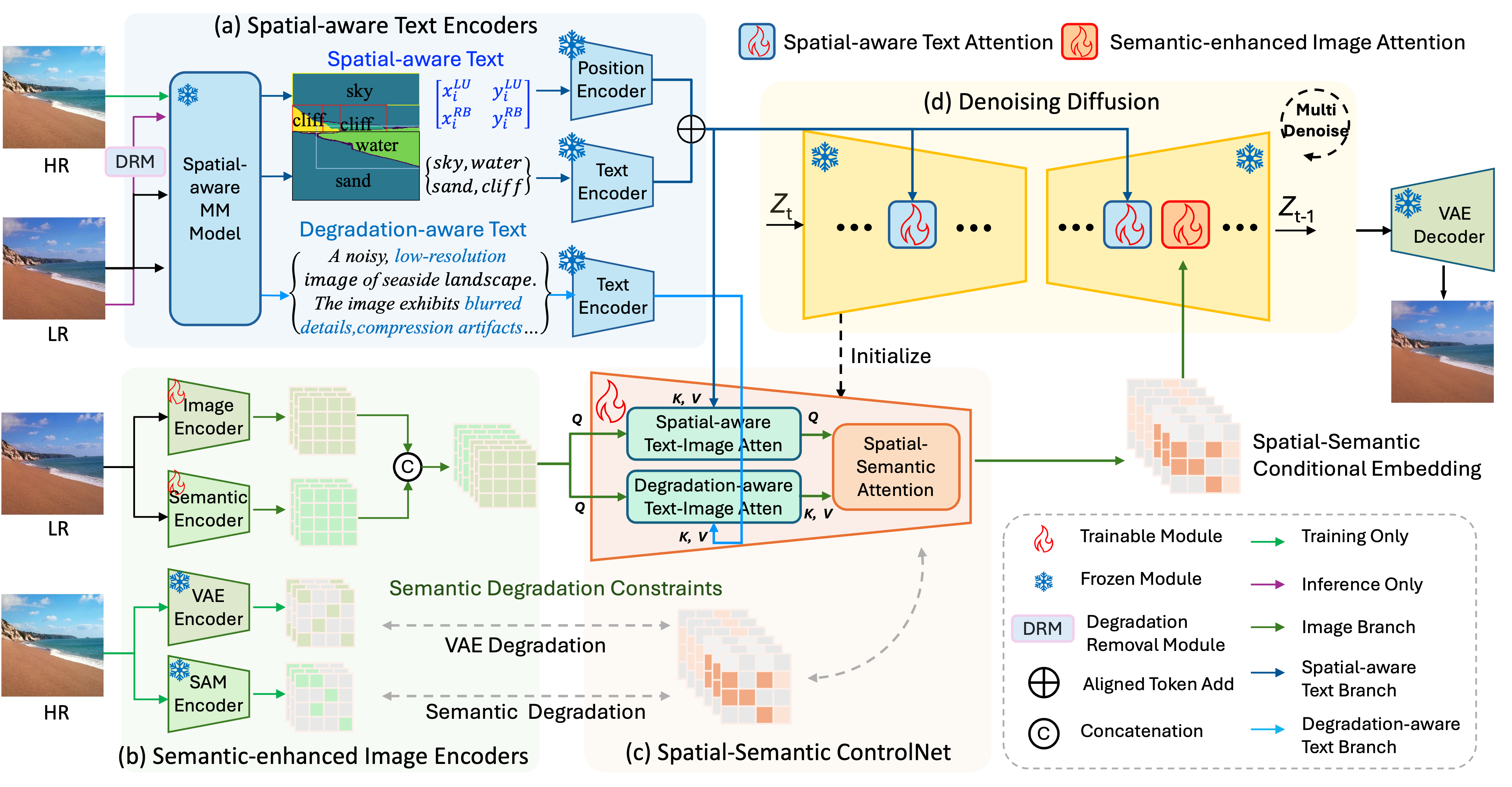} 
    \caption{Framework overview. (a) Spatial-aware text encoders generate position-grounded textual prompts (Sec. \ref{Sec:Spatial-awareTextEncoders}); (b) Semantic-enhanced image encoders extract semantic-enhanced visual features with degradation constraints (Sec. \ref{sec:Semantic-enhancedImageEncoders}); (c) Spatial-semantic ControlNet integrates these multimodal conditions (Sec. \ref{sec:Spatial-SemanticControlNet}); (d) Spatial-semantic guided diffusion fuses semantic and spatial guidance with generative priors (Sec. \ref{sec:Spatial-SemanticDiffusion}).}
    \label{fig:architecture}
\end{figure*}

\textbf{Preliminary: Stable Diffusion.}
\label{Sec:preliminary}
Our method builds on Stable Diffusion (SD), a latent diffusion model for T2I generation. SD operates in a compressed latent space for efficiency, where an autoencoder maps an image to latent $\mathbf{z}_0 = \mathcal{E}(I_0)$ and reconstructs it as $I_0 = \mathcal{D}(\mathbf{z}_0)$.  

The forward diffusion gradually perturbs $\mathbf{z}_0$ with Gaussian noise:
\begin{equation}
q(\mathbf{z}_t \mid \mathbf{z}_{t-1}) = \mathcal{N}(\mathbf{z}_t; \sqrt{1-\beta_t}\mathbf{z}_{t-1}, \beta_t \mathbf{I}).
\end{equation}

The reverse process recovers $\mathbf{z}_{t-1}$ from $\mathbf{z}_t$ via a denoising network $\epsilon_\theta$:
\begin{equation}
p_\theta(\mathbf{z}_{t-1} \mid \mathbf{z}_t) = \mathcal{N}(\mathbf{\mu}_\theta(\mathbf{z}_t, t), \mathbf{\Sigma}_\theta(\mathbf{z}_t, t)),
\end{equation}
which is trained to minimize the noise prediction objective:
\begin{equation}
\mathcal{L}_{SD} = \mathbb{E}_{\mathbf{z}_0, t, \epsilon}\Big[\|\epsilon - \epsilon_\theta(\sqrt{\bar{\alpha}_t}\mathbf{z}_0 + \sqrt{1-\bar{\alpha}_t}\epsilon, t)\|_2^2\Big],
\end{equation}
where $\epsilon \sim \mathcal{N}(0, \mathbf{I})$, $\bar{\alpha}_t = \prod_{i=0}^{t} (1-\beta_i)$, $t=1,\dots,T$. Generation starts from noise $\mathbf{z}_T \sim \mathcal{N}(0, \mathbf{I})$ and iteratively applies reverse denoising to obtain $\mathbf{z}_0$.

\textbf{Motivation.}
Image SR faces the well-known perception-distortion trade-off~\cite{blau2018perception}: GAN-based methods~\cite{zhang2021designing,wang2021real} achieve strong fidelity but often yield over-smoothed results lacking realistic details, while diffusion-based methods~\cite{wang2024exploiting,lin2024diffbir} generate sharper details and higher perceptual quality but frequently hallucinate content, leading to fidelity loss. This motivates our goal: harnessing the generative strength of diffusion while constraining it with spatial and semantic priors to reduce distortion. 


\textbf{Framework Overview.}
We propose SpaSemSR, a Spatial-Semantic guided SR framework that steers diffusion with two complementary forms of guidance (Fig.~\ref{fig:architecture}):  
\textbf{(a)} \textit{Spatial-grounded textual guidance}, which integrates semantic prompts with object-level spatial cues, improving fidelity by aligning text semantics with visual structure (Sec.~\ref{Sec:Spatial-awareTextEncoders});  
\textbf{(b)}\textit{ Semantic-enhanced visual guidance}, which extracts semantic-enhanced features from degraded LR inputs under semantic degradation constraints, improving perceptual realism (Sec.~\ref{sec:Semantic-enhancedImageEncoders}).  
These guidances are fused in \textbf{(c)} the proposed \emph{Spatial-Semantic ControlNet} (Sec.~\ref{sec:Spatial-SemanticControlNet}) via parallel cross-attention and integrated into \textbf{(d)} our \textit{Spatial-Semantic guided Diffusion model} (Sec.~\ref{sec:Spatial-SemanticDiffusion}), where they adaptively modulate the generative prior. By integrating spatial-grounded textual and semantic-enhanced visual guidance, SpaSemSR achieves reconstructions that are both perceptually realistic and faithful to the ground truth, effectively balancing the perception-distortion trade-off.


\subsection{Spatial-aware Text Encoders}
\label{Sec:Spatial-awareTextEncoders}
In conventional T2I models, text inputs guide diffusion models but lack spatial grounding. As a result, textual prompts may misalign semantics with image regions, leading to reduced fidelity and unwanted distortions. To address this, we propose spatial-aware text encoders that integrate both spatially grounded semantics and degradation-aware information.  

\textbf{Spatial-grounded Text Representations.}
We introduce a spatial-aware text fusion mechanism that combines object semantics with their spatial locations, yielding more faithful region-to-text alignment. Specifically, given a low-resolution image $I_{LR}$, we employ a pretrained Grounded-SAM~\cite{ren2024grounded} to extract object-level textual tags $\mathbf{x}_\text{obj-text}$ and corresponding bounding boxes $\mathbf{x}_\text{spa}$: 
\begin{equation}
\begin{aligned}
\mathbf{x}_\text{obj-text} = \mathcal{F}_\text{obj-text}(I_{LR}),\quad
\mathbf{x}_\text{spa} = \mathcal{F}_\text{spa}(I_{LR}),
\end{aligned}
\end{equation}
where $\mathcal{F}_\text{obj-text}$ and $\mathcal{F}_\text{spa}$ are the object recognition and bounding box model from the pretrained Grounded-SAM. 

The spatial coordinates are encoded with sinusoidal positional encoding $\mathcal{F}_\text{PE}(\cdot)$ to obtain embeddings $\mathbf{e}_\text{spa}$, which are then fused object-wise with textual embeddings $\mathbf{e}_\text{obj-text}$ from the pretrained CLIP encoder \( \mathcal{E}_\text{text}\) ~\cite{radford2021learning}:
\begin{equation}
\begin{aligned}
&\mathbf{e}_\text{spa} = \mathcal{F}_\text{PE}(\mathbf{x}_\text{spa}), \\
&\mathbf{e}_\text{obj-text} = \mathcal{E}_\text{text}(\mathbf{x}_\text{obj-text}), \\
&\mathbf{e}_\text{spa-text} = \mathcal{F}_\text{fusion}(\mathbf{e}_\text{obj-text},\mathbf{e}_\text{spa}),
\end{aligned}
\end{equation}
where $\mathcal{F}_\text{fusion}(\cdot)$ is the object-wise fusion function. This ensures a unique semantic-spatial correspondence, producing spatial-grounded text embeddings $\mathbf{e}_\text{spa-text}$ that better preserve fidelity.  

\textbf{Degradation-aware Text Representations.}
Object-level tags capture semantics but not degradation characteristics (e.g., blur, noise, compression). To complement them, we adopt degradation-aware textual priors~\cite{qu2024xpsr} by using LLaVA~\cite{liu2023visual} to generate an image-level description $\mathbf{x}_\text{deg-text}$ that encodes coarse attributes such as sharpness and noise:
\begin{equation}
\begin{aligned}
\mathbf{x}_\text{deg-text} = \mathcal{F}_\text{deg-text}(I_{LR}),\quad
\mathbf{e}_\text{deg-text}= \mathcal{E}_\text{text}(\mathbf{x}_\text{{deg-text}}),
\end{aligned}
\end{equation}
where $\mathcal{F}_\text{deg-text}$ is the pretrained degradation-aware textual generation model. For each $I_{LR}$, we obtain two complementary textual priors:  (i) spatial-grounded semantic prompts $\mathbf{e}_\text{spa-text}$ that enhance region-level semantics, and (ii) degradation-aware prompts $\mathbf{e}_\text{deg-text}$ that facilitate degradation modeling. Finally, these dual embeddings are jointly fed into ControlNet and the diffusion model, enabling high-fidelity, spatially consistent, and degradation-aware textual representations.  

\subsection{Semantic-enhanced Image Encoders} 
\label{sec:Semantic-enhancedImageEncoders}
Image SR aims to reconstruct an HR image from a degraded LR input. Diffusion-based SR models typically rely on LR images as control conditions. However, severe degradation often destroys local structures, leading to ambiguous or misleading semantics.  

To address this, we design a dual-encoder system with our semantic degradation loss that extracts complementary features from the LR input: (1) a low-level encoder $\mathcal{E}_\text{img}$ that captures latent structural details, and (2) a high-level semantic encoder $\mathcal{E}_\text{sem}$ that emphasizes semantic consistency. The perception ability of these features is further reinforced by semantic degradation constraints derived from pretrained VAE~\cite{rombach2022high} and SAM~\cite{kirillov2023segment} encoders. This design produces semantic-enhanced features that are robust to degradation and preserve meaningful visual semantics. 
Formally, given an LR image $I_{LR}$, the two encoders extract high-level semantic features $\mathbf{x}_\text{sem}$ and low-level latent features $\mathbf{x}_\text{img}$, respectively:

\begin{equation}
\begin{aligned}
&\mathbf{x}_\text{img} = \mathcal{E}_\text{img}(I_{LR}),\\
&\mathbf{x}_\text{sem} = \mathcal{E}_\text{sem}(I_{LR}),\\
&\mathbf{x}_\text{sem-img} = \text{Concat}(\mathbf{x}_\text{img}, \mathbf{x}_\text{sem}), 
\end{aligned}
\end{equation}
where $\text{Concat}(\cdot)$ denotes the concatenation operation. The fused feature $\mathbf{x}_\text{sem-img}$ is injected into each ControlNet layer. Following~\cite{qu2024xpsr}, we extract a hybrid representation $\mathbf{x}_\text{hyb}^i$ from the $i$-th layer and evenly split it into two streams:

\begin{equation}
[\mathbf{x}_\text{img}^i, \mathbf{x}_\text{sem}^i] = \text{Split}(\mathbf{x}_\text{hyb}^i),
\end{equation}
where $\mathbf{x}_\text{img}^i$ and $\mathbf{x}_\text{sem}^i$ retain the respective low-level and semantic branches, ensuring disentangled feature learning.

\paragraph{Semantic Degradation Loss.}
To guide each encoder, we introduce a semantic degradation loss:
\begin{equation}
\begin{aligned}
\mathcal{L}^\text{SemDeg}_\text{ControlNet} =&\sum_{i=1}^{n}[(1 - \lambda)\|\mathbf{x}_\text{img}^i - \hat{\mathbf{x}}_\text{img}^i\|_1 \\
&+ \lambda\|\mathbf{x}_\text{sem}^i - \hat{\mathbf{x}}_\text{sem}^i\|_1],
\end{aligned}
\end{equation}
where $\hat{\mathbf{x}}_\text{img}^i$ and $\hat{\mathbf{x}}_\text{sem}^i$ are the reference features extracted from pretrained VAE and SAM encoders on the HR image, respectively. The balancing coefficient $\lambda \in [0,1]$ controls the relative emphasis on structural fidelity versus semantic consistency. By enforcing this loss, each encoder is encouraged to specialize: low-level encoders align with structural details, while semantic encoders capture high-level semantics, thereby providing more reliable features to guide the diffusion model under complex degradations.

\subsection{Spatial-Semantic ControlNet}
\label{sec:Spatial-SemanticControlNet}
To fully leverage semantic-enhanced image priors and spatial-grounded textual knowledge, we design a Spatial-Semantic ControlNet, which serves as a controller integrated with Stable Diffusion. 

Specifically, we introduce two parallel and one fusion attention modules: \textit{Spatial-aware Text-Image Attention} (SpaAtten) and \textit{Degradation-aware Text-Image Attention} (DegAtten), followed by a \textit{Spatial-Semantic Attention} (SpaSemAtten) fusion layer. As shown in Fig.~\ref{fig:architecture}, the spatial-aware text features $\mathbf{e}_\text{spa-text}$ and degradation-aware text features $\mathbf{e}_\text{deg-text}$, obtained from Sec.~\ref{Sec:Spatial-awareTextEncoders}, serve as the key-value sets for SpaAtten and DegAtten, respectively. The semantic-enhanced image features $\mathbf{x}_\text{sem-img}$ (Sec.~\ref{sec:Semantic-enhancedImageEncoders}) are shared as queries across both branches. The outputs of SpaAtten and DegAtten are then fused by SpaSemAtten to integrate complementary cues. Formally, the attention operations are defined as:
\begin{equation}
\begin{aligned}
&\mathbf{y}_\text{spa} = \text{SpaAtten}(\mathbf{x}_\text{sem-img}, \mathbf{e}_\text{spa-text}, \mathbf{e}_\text{spa-text}), \\
&\mathbf{y}_\text{deg} = \text{DegAtten}(\mathbf{x}_\text{sem-img}, \mathbf{e}_\text{deg-text}, \mathbf{e}_\text{deg-text}), \\
&\mathbf{x}_\text{spa-sem} = \text{SpaSemAtten}(\mathbf{y}_\text{spa}, \mathbf{y}_\text{deg}, \mathbf{y}_\text{deg}),
\end{aligned}
\end{equation}
where $\mathbf{y}_\text{spa}$ and $\mathbf{y}_\text{deg}$ are intermediate outputs from the two branches, and $\mathbf{x}_\text{spa-sem}$ denotes the fused representation. Each cross-attention follows the standard formulation: \(\text{Atten}(Q,K,V) = \text{Softmax}\!\left(\frac{QK^{T}}{\sqrt{d}}\right)V.\)
This design allows SpaAtten to inject precise spatial textual cues while DegAtten introduces degradation-aware textual context. The subsequent SpaSemAtten layer fuses both signals, balancing local spatial fidelity with global semantic consistency. Therefore, the diffusion model generates contents that are not only perceptually realistic but also structurally faithful. 

\subsection{Diffusion via Spatial-Semantic Guidance}
\label{sec:Spatial-SemanticDiffusion}
To seamlessly integrate cross-modal semantic and spatial priors with T2I generative priors, we introduce the spatial-semantic guided diffusion model (\textit{SpaSemDM}). SpaSemDM adaptively learns to integrate semantic and spatial cues into the denoising process, thereby producing reconstructions that are both perceptually compelling and faithful to the source content.

In SpaSemDM, the denoising network is augmented with two control conditions: (1) spatially aware textual features $\mathbf{e}_\text{spa-text}$ extracted from Spatial-Aware Text Encoders, and (2) spatial-semantic visual embeddings $\mathbf{x}_\text{spa-sem}$ from the Spatial-Semantic ControlNet. To fuse these conditions, we design two additional attention modules: \textit{Spatial-aware Text Attention} (SpaTextAtten) and \textit{Semantic-enhanced Image Attention} (SemImgAtten), which jointly inject spatial grounding and semantic alignment into the latent space during diffusion learning. This mechanism allows SpaSemDM to better navigate the perception-distortion trade-off.

\textbf{Training and Optimization.} During training, we first obtain the latent representation $z_{0}$ of an HR image, which is progressively corrupted by Gaussian noise to yield $z_{t}$ at step $t$. Conditioned on $t$, the LR input $I_{\mathrm{LR}}$, its degradation-aware text prompt $\mathbf{x}_\text{deg-text}$, and the spatial-grounded text prompt $\{\mathbf{x}_\text{obj-text}, \mathbf{x}_\text{spa}\}$, SpaSemDM network $\epsilon_{\theta}$ is trained to predict the noise added to $z_{t}$. The objective is:
\begin{equation}
\begin{aligned}
\mathcal{L}_{SD}^{\text{SpaSem}}
=& \mathbb{E}_{z_0, t, I_{\mathrm{LR}}, \epsilon} \Big[
  \big\lVert 
    \epsilon -
    \epsilon_{\theta}\big(
      \mathbf{z}_t, t, \,
      I_{\mathrm{LR}},
      \mathbf{x}_{\text{deg-text}},
\\[-2pt]
&\{\mathbf{x}_{\text{obj-text}},
      \mathbf{x}_{\text{spa}}\}
    \big)
  \big\rVert_2^2
\Big].
\end{aligned}
\end{equation}


 In our model training, the final loss can be expressed as:
 \begin{equation}
\mathcal{L} 
= \mathcal{L}_{SD}^\text{SpaSem}+\lambda_\text{ControlNet}\mathcal{L}^\text{SemDeg}_\text{ControlNet},
\end{equation}
where $\mathcal{L}^\text{SemDeg}_\text{ControlNet}$ regularizes the image encoders and ControlNet to enhance semantic fidelity, and  $\lambda_\text{ControlNet}$ is the weighting coefficient.

\section{Experiments}
\label{sec:expt}

\begin{figure*}[t]
  \centering
    \includegraphics[width=0.9\textwidth]{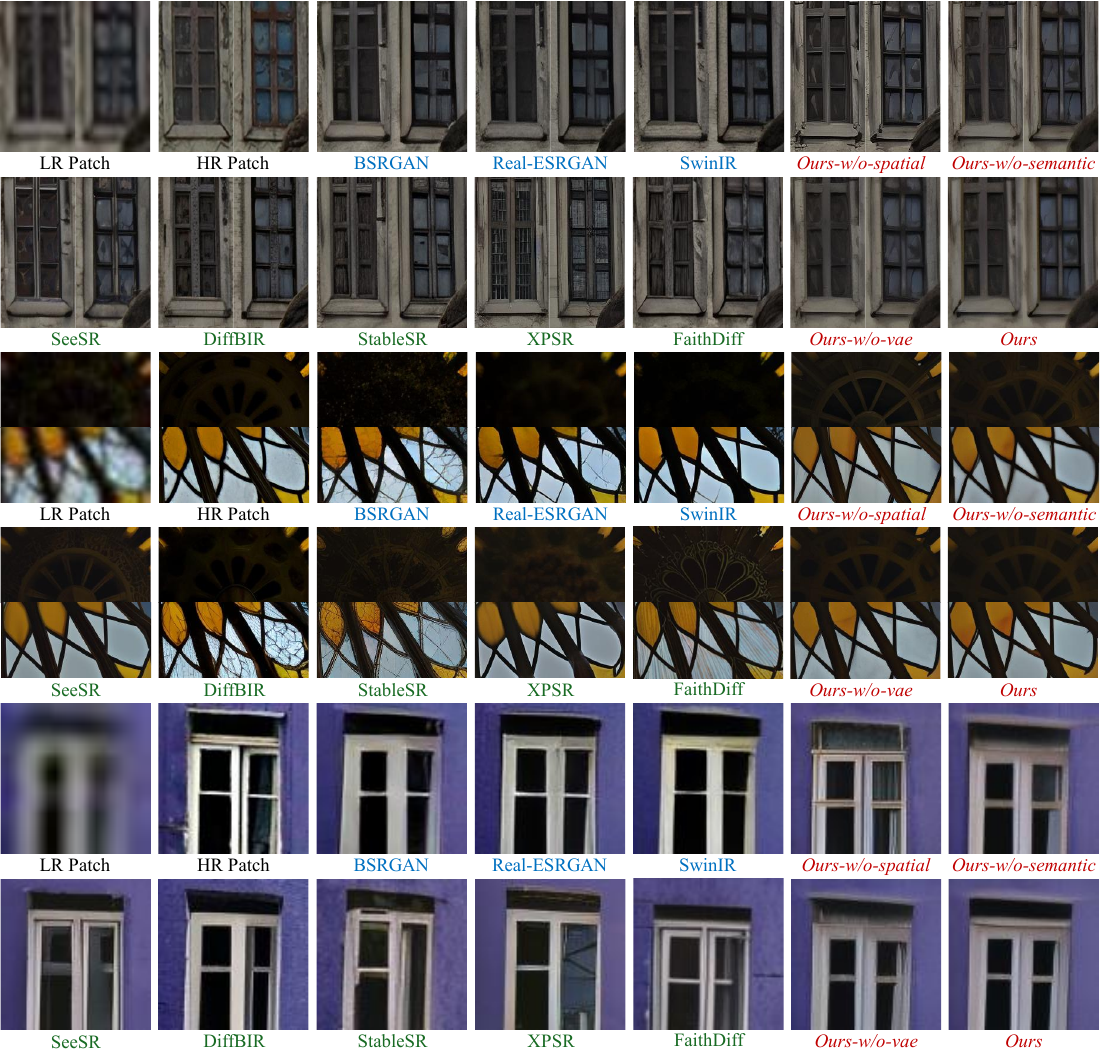}
  \caption{Qualitative comparisons with different methods. Zoom in for a better view.}
  \label{fig:Example2}
\end{figure*}

\begin{table}[t]
\centering
\caption{Comparison with GAN methods. \textcolor{red}{Red} and \textcolor{blue}{blue} indicate the best and second-best results.}
\label{tab:gan_pivot}
\resizebox{0.9\columnwidth}{!}{
\begin{tabular}{c | c | *{2}{c} | *{3}{c} }
\toprule
\multirow{2}{*}{Dataset} 
& \multirow{2}{*}{Model}
& \multicolumn{2}{c|}{Reference Fidelity}
& \multicolumn{3}{c}{Non-reference Perception}\\
\cmidrule(lr){3-4} \cmidrule(lr){5-7}
& 
& PSNR$\uparrow$
& SSIM$\uparrow$
& CLIP-IQA$\uparrow$
& MUSIQ$\uparrow$
& MANIQA$\uparrow$ \\
\midrule
\multirow{4}{*}{\makecell{DIV2K-\\Val}}
& BSRGAN  & \textcolor{blue}{21.74} & 0.5530 & 0.5234 & \textcolor{blue}{59.16} & 0.3528 \\
& \makecell{Real-\\ESRGAN} & \textcolor{red}{21.86} & \textcolor{red}{0.5746} & \textcolor{blue}{0.5485} & 58.80 & \textcolor{blue}{0.3776} \\
& SwinIR & 21.45 & \textcolor{blue}{0.5639} & 0.5467 & 59.03 & 0.3634 \\
\cmidrule(lr){2-7}
& \cellcolor{gray!15}\textbf{Ours} & \cellcolor{gray!15}21.31 & \cellcolor{gray!15}0.5340 & \cellcolor{gray!15}\textcolor{red}{0.6932} & \cellcolor{gray!15}\textcolor{red}{63.32} & \cellcolor{gray!15}\textcolor{red}{0.4945} \\
\midrule
\multirow{4}{*}{RealSR}
& BSRGAN & \textcolor{red}{26.38} & \textcolor{blue}{0.7651} & \textcolor{blue}{0.5112} & \textcolor{red}{63.28} & \textcolor{blue}{0.3754} \\
& \makecell{Real-\\ESRGAN} & 25.69 & 0.7614 & 0.4490 & 60.37 & 0.3730 \\
& SwinIR & \textcolor{blue}{26.31} & \textcolor{red}{0.7729} & 0.4364 & 58.69 & 0.3444 \\
\cmidrule(lr){2-7}
& \cellcolor{gray!15}\textbf{Ours} & \cellcolor{gray!15}25.74 & \cellcolor{gray!15}0.7306 & \cellcolor{gray!15}\textcolor{red}{0.5893} & \cellcolor{gray!15}\textcolor{blue}{62.96} & \cellcolor{gray!15}\textcolor{red}{0.4637} \\
\midrule
\multirow{4}{*}{DrealSR}
& BSRGAN  & \textcolor{blue}{28.70} & 0.8028 & \textcolor{blue}{0.5093} & \textcolor{red}{57.16} & 0.3447 \\
& \makecell{Real-\\ESRGAN} & 28.61 & \textcolor{red}{0.8052} & 0.4517 & 54.27 & \textcolor{blue}{0.3448} \\
& SwinIR  & 28.50 & \textcolor{blue}{0.8044} & 0.4445 & 52.74 & 0.3298 \\
\cmidrule(lr){2-7}
& \cellcolor{gray!15}\textbf{Ours}  & \cellcolor{gray!15}\textcolor{red}{28.97} & \cellcolor{gray!15}0.7826 & \cellcolor{gray!15}\textcolor{red}{0.5630} & \cellcolor{gray!15}\textcolor{blue}{55.79} & \cellcolor{gray!15}\textcolor{red}{0.3998} \\
\bottomrule
\end{tabular}}
\end{table}

\begin{table}[t]
\centering
\caption{Comparison with diffusion methods. \textcolor{red}{Red} and \textcolor{blue}{blue} indicate the best and second-best results.}
\label{tab:diffusion_pivot}
\resizebox{0.9\columnwidth}{!}{
\begin{tabular}{c | c | *{2}{c} | *{3}{c} }
\toprule
\multirow{2}{*}{Dataset} & \multirow{2}{*}{Model}
  & \multicolumn{2}{c|}{Reference Fidelity}
  & \multicolumn{3}{c}{Non-reference Perception} \\
\cmidrule(lr){3-4} \cmidrule(lr){5-7}
 & 
  & PSNR$\uparrow$ & SSIM$\uparrow$ 
  & CLIP-IQA$\uparrow$ & MUSIQ$\uparrow$ & MANIQA$\uparrow$ \\
\midrule
\multirow{7}{*}{\makecell{DIV2K-\\Val}}
& StableSR  & 20.74 & 0.4888 & 0.6605 & 63.19 & 0.4002 \\
& DiffBIR   & 20.57 & 0.4740 & \textcolor{blue}{0.7359} & \textcolor{blue}{69.93} & \textcolor{blue}{0.5763} \\
& PASD      & 20.77 & 0.5022 & 0.6140 & 63.29 & 0.4581 \\
& SeeSR     & \textcolor{blue}{21.00} & \textcolor{red}{0.5362} & 0.7074 & 68.81 & 0.5149 \\
& XPSR      & 20.56 & 0.5081 & \textcolor{red}{0.7826} & \textcolor{red}{70.07} & \textcolor{red}{0.6108} \\
& FaithDiff & 20.63 & 0.4999 & 0.6545 & 69.62 & 0.4427 \\
\cmidrule(lr){2-7}
& \cellcolor{gray!15}\textbf{Ours} & \cellcolor{gray!15}\textcolor{red}{21.31} & \cellcolor{gray!15}\textcolor{blue}{0.5340} & \cellcolor{gray!15}0.6932 & \cellcolor{gray!15}63.32 & \cellcolor{gray!15}0.4945 \\
\midrule
\multirow{7}{*}{RealSR}
& StableSR  & 24.70 & 0.7085 & 0.6166 & 65.18 & 0.4178 \\
& DiffBIR   & 24.83 & 0.6501 & \textcolor{blue}{0.7054} & 69.28 & \textcolor{blue}{0.5596} \\
& PASD      & \textcolor{blue}{25.26} & 0.7191 & 0.6249 & 67.78 & 0.4971 \\
& SeeSR     & 25.15 & \textcolor{blue}{0.7210} & 0.6704 & \textcolor{blue}{69.82} & 0.5395 \\
& XPSR      & 23.74 & 0.6734 & \textcolor{red}{0.7417} & \textcolor{red}{71.45} & \textcolor{red}{0.6293} \\
& FaithDiff & 25.25 & 0.7054 & 0.6152 & 68.78 & 0.4628 \\
\cmidrule(lr){2-7}
& \cellcolor{gray!15}\textbf{Ours} & \cellcolor{gray!15}\textcolor{red}{25.74} & \cellcolor{gray!15}\textcolor{red}{0.7306} & \cellcolor{gray!15}0.5893 & \cellcolor{gray!15}62.96 & \cellcolor{gray!15}0.4637 \\
\midrule
\multirow{7}{*}{DrealSR}
& StableSR  & \textcolor{blue}{28.07} & 0.7489 & 0.6375 & 58.99 & 0.3892 \\
& DiffBIR   & 25.90 & 0.6245 & \textcolor{blue}{0.7068} & 66.13 & \textcolor{blue}{0.5526} \\
& PASD      & 27.07 & 0.7251 & 0.6710 & 64.56 & 0.5061 \\
& SeeSR     & \textcolor{blue}{28.07} & \textcolor{blue}{0.7684} & 0.6911 & 65.09 & 0.5115 \\
& XPSR      & 26.55 & 0.7289 & \textcolor{red}{0.7433} & \textcolor{red}{67.02} & \textcolor{red}{0.5684} \\
& FaithDiff & 27.42 & 0.7193 & 0.6352 & \textcolor{blue}{66.58} & 0.4502 \\
\cmidrule(lr){2-7}
& \cellcolor{gray!15}\textbf{Ours} & \cellcolor{gray!15}\textcolor{red}{28.97} & \cellcolor{gray!15}\textcolor{red}{0.7826} & \cellcolor{gray!15}0.5630 & \cellcolor{gray!15}55.79 & \cellcolor{gray!15}0.3998 \\
\bottomrule
\end{tabular}}
\end{table}

\begin{figure*}[t]
  \centering
    \includegraphics[width=0.9\textwidth]{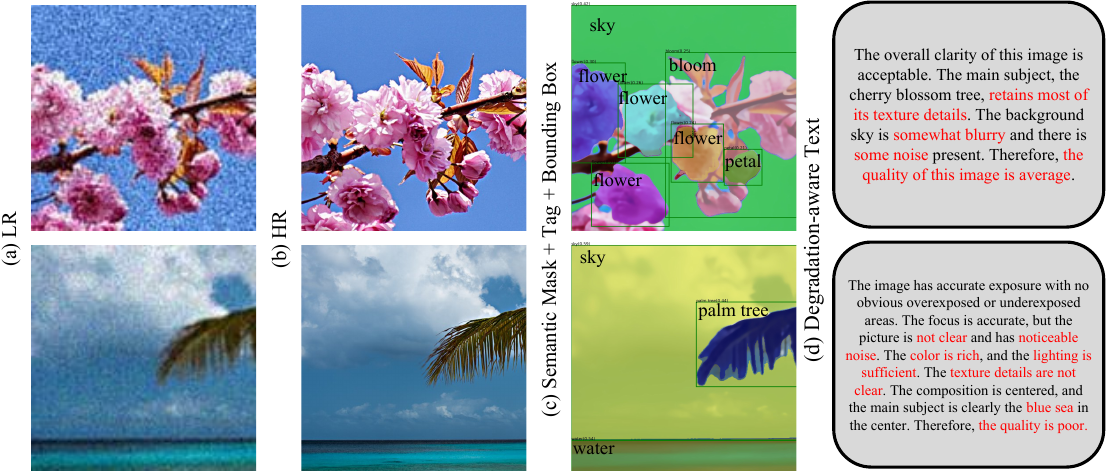}
\caption{Visualization of spatial-grounded textual and semantic-enhanced visual guidance.}
  \label{fig:guidance}
\end{figure*}

\textbf{Implementation.} Our framework is built on ControlNet~\cite{zhang2023adding} with Stable Diffusion v2~\cite{rombach2022high} as the backbone. Semantic features from a pretrained SAM image encoder and image features from a VAE encoder serve as constraints for the corresponding encoders. During training, Grounded-SAM~\cite{ren2024grounded} provides positional information and object-level tags from HR images. For training inference consistency, LR inputs in inference are processed by a degradation removal module (DRM) to restore a clean image following~\cite{chen2025faithdiff} , and then bounding boxes are extracted from this restored image with Grounded-SAM. Training runs for 200k iterations (batch size 32, lr \(5 \times 10^{-5}\)) at \(512 \times 512\) resolution on 4×RTX 6000 GPUs for three days. 

\textbf{Datasets and Metrics.} \textit{Training Data:} We train on DIV2K~\cite{agustsson2017ntire}, DIV8K~\cite{gu2019div8k}, Flickr2K~\cite{timofte2017ntire}, OutdoorSceneTraining~\cite{kim2021noise}, Unsplash2K~\cite{wang2018recovering}, and 5k FFHQ faces~\cite{karras2019style}, using Real-ESRGAN’s~\cite{wang2021real} degradation pipeline to synthesize LR-HR pairs.
\textit{Test Data:} Synthetic evaluation uses 3,000 degraded DIV2K validation patches with the same pipeline as Real-ESRGAN. For real-world data, we follow StableSR~\cite{wang2024exploiting}, evaluate on RealSR~\cite{cai2019toward} and DRealSR~\cite{wei2020component}, center-cropped to 128×128 LR images. We also evaluate on RealLR200~\cite{wu2024seesr}, which lacks ground-truth.
\textit{Metrics:} For distortion, we report PSNR and SSIM~\cite{wang2004image}. For perceptual quality, we adopt CLIP-IQA~\cite{wang2023exploring}, MUSIQ~\cite{ke2021musiq}, and MANIQA~\cite{yang2022maniqa}.

\subsection{Balancing Perception and Distortion}
A key challenge in SR is reconciling the perception-distortion trade-off. We evaluate our SpaSemSR method against representative GAN-based methods (Real-ESRGAN~\cite{wang2021real}, BSRGAN~\cite{zhang2021designing}, SwinIR~\cite{liang2021swinir}) and diffusion-based methods (StableSR~\cite{wang2024exploiting}, DiffBIR~\cite{lin2024diffbir}, PASD~\cite{yang2024pixel}, SeeSR~\cite{wu2024seesr}, XPSR~\cite{qu2024xpsr}, and FaithDiff~\cite{chen2025faithdiff}).

\textbf{Quantitative Comparisons.}

\textbf{(1) Perceptual Quality: }
As shown in Table~\ref{tab:gan_pivot}, our model significantly outperforms GAN-based methods on CLIP-IQA and MANIQA, and consistently achieves higher MUSIQ scores. For example, on the DIV2K-Val dataset, our method surpasses the second-best result (highlighted in blue) by 26.4\%, 7.03\%, and 31.0\% in CLIP-IQA, MUSIQ, and MANIQA, respectively. We attribute this gain primarily to the strong generative capability of the diffusion model, as well as the incorporation of semantic-enhanced guidance, which facilitates more realistic and semantically consistent detail synthesis.
\textbf{(2) Reconstruction Fidelity: }
Table~\ref{tab:diffusion_pivot} indicates that our model outperforms diffusion-based methods in terms of reconstruction fidelity. Our model exhibits only a minor gap compared with GAN-based methods and even surpass them on the DRealSR real-world images in terms of PSNR in Table~\ref{tab:gan_pivot}. These results demonstrate that our model can effectively enhance the fidelity of diffusion-based models and achieve high fidelity comparable to GAN-based methods.
\textbf{(3) Trade-off analysis: } GAN-based methods often suffer from overly smooth textures and blurry artifacts, resulting in limited perceptual quality. Diffusion-based methods tend to generate visually appealing details but frequently introduce hallucinated details that are inconsistent with the ground truth, thereby sacrificing reconstruction fidelity. Our model substantially enhances perceptual quality compared with GAN-based methods and effectively elevates the upper bound of reconstruction fidelity compared with diffusion-based methods. This complementary evaluation strategy more clearly demonstrates how our model bridges the gap between the two paradigms. Our approach balances the quantitative behavior of GAN-based and diffusion-based methods while mitigating their respective weaknesses, thereby achieving a better perception-distortion trade-off.

\textbf{Qualitative Comparisons.}
Visual comparisons (Fig.~\ref{fig:Example2}) further validate our approach. 
\textbf{(1) Perceptual Quality: } GAN-based methods typically produce over-smoothed or blurry reconstructions that lack human-perceived realism. In the case of windows from Fig.~\ref{fig:Example2}, they struggle to generate plausible window frame textures, resulting in overly-smoothed or blurry reconstructions that fall short in texture richness and sharpness compared to diffusion models, resulting in inferior perceptual quality. In contrast, our four variants reconstruct clearer window-frame textures and sharper, perceptually plausible details, yielding higher perceived quality than GAN-based methods. \textbf{(2) Reconstruction Fidelity: } Although diffusion-based methods can generate clear textures and details, their inherent generative diversity often results in details and textures inconsistent with the ground truth. For example, SeeSR and DiffBIR produce inconsistent window frame textures, while StableSR, FaithDiff, and XPSR generate extra curtains or window frames on originally reflective window glass. This diversity-driven hallucination undermines fidelity, causes semantic misalignment with the ground truth. In comparison, our model produces images with realistic, fine-grained textures that closely match the ground truth, exhibiting minimal distortion and substantially fewer hallucinated artifacts. \textbf{(3) Trade-off analysis: } Our methods deliver clearer, fine-grained textures than GAN-based methods, achieve higher perceptual quality, and exhibit fewer hallucinations with better ground truth alignment than diffusion-based methods, producing high-fidelity reconstructions with a well-balanced perception-distortion trade-off. More qualitative results are provided in the supplementary material.

\subsection{Ablation Study}
\label{sec:ablation}

To validate the contributions of the proposed \emph{spatial-grounded textual guidance} and \emph{semantic-enhanced visual guidance}, we construct several model variants for ablation analysis: (i) \textbf{Ours-w/o-semantic}: retains only spatial-grounded textual guidance with image-encoder constraints, without semantic-encoder constraints; (ii) \textbf{Ours-w/o-spatial}: employs semantic-enhanced visual guidance (image and semantic encoders with constraints), without spatial grounding; (iii) \textbf{Ours-w/o-vae}: combines spatial-grounded textual guidance with semantic-encoder constraints; (iv) \textbf{Ours}: full model combining spatial-grounded textual guidance with both image and semantic encoder constraints. Fig.~\ref{fig:guidance} displays the content of spatial-grounded textual and semantic-enhanced visual guidance.  Table~\ref {tab:perception_balance_compare} and Table~\ref {tab:fidelity_balance_compare} separately report the results of semantic-enhanced visual guidance and spatial-grounded textual guidance on synthetic and real-world datasets.

\textbf{Guidance visualization.}
In Fig.~\ref{fig:guidance}(c) and (d), the object-level tags with their corresponding bounding boxes, together with the degradation-aware text, constitute the spatial-grounded textual guidance. And the semantic mask in Fig.~\ref{fig:guidance}(c) serves as prior information of our semantic-enhanced visual guidance. Incorporating high-level semantic priors extracted from the SAM image encoder as guidance contributes to improving perceptual quality. In addition, guiding the model to align the visual semantic mask with the corresponding object-level tags further enhances image fidelity, leading to a better trade-off between perceptual quality and distortion. 

\begin{figure*}[t]
  \centering
    \includegraphics[width=0.9\textwidth]{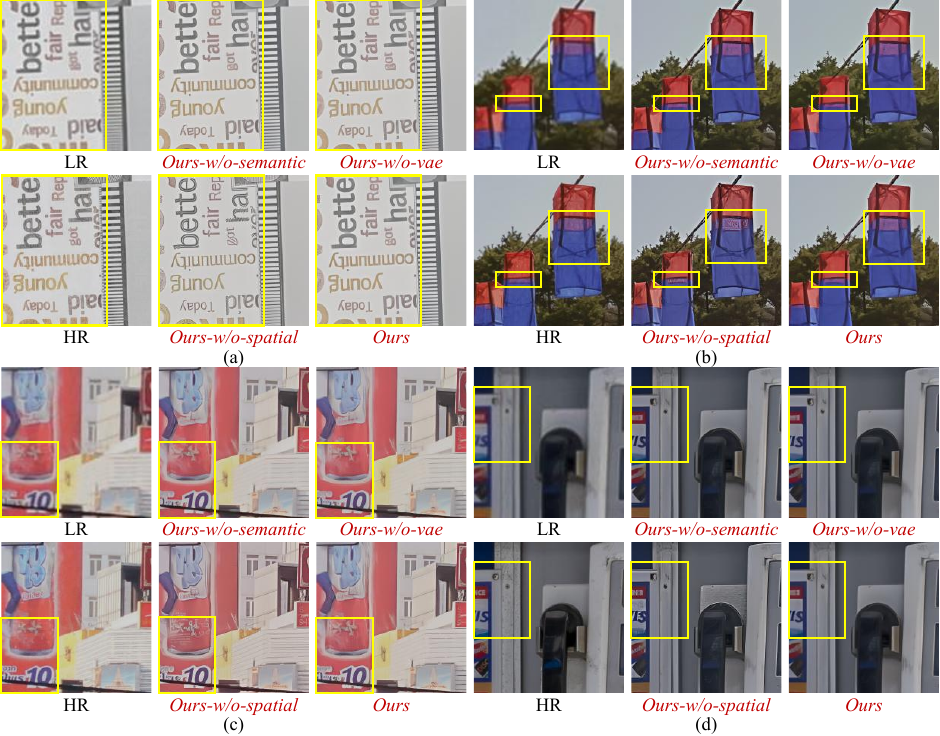}
\caption{Ablation visualization with different variants on real-world datasets. Zoom in for a better view.}
\label{fig:Ablation_vis}
\end{figure*}

\begin{table*}[h]
\centering
\caption{Ablation on semantic-enhanced visual guidance.}
\label{tab:perception_balance_compare}
\resizebox{0.9\linewidth}{!}{
\begin{tabular}{c|ccc|ccc|ccc}
\toprule
 & \multicolumn{3}{c|}{RealSR} 
 & \multicolumn{3}{c|}{DRealSR}
 & \multicolumn{3}{c}{DIV2K-Val} \\
\cmidrule(lr){2-4} \cmidrule(lr){5-7} \cmidrule(lr){8-10}
Method
& CLIP-IQA$\uparrow$ & MUSIQ$\uparrow$ & MANIQA$\uparrow$
& CLIP-IQA$\uparrow$ & MUSIQ$\uparrow$ & MANIQA$\uparrow$
& CLIP-IQA$\uparrow$ & MUSIQ$\uparrow$ & MANIQA$\uparrow$ \\
\midrule
Ours-w/o-semantic 
& 0.5683 & 61.82 & 0.4332
& 0.5538 & 54.31 & 0.3829
& 0.6523 & 61.88 & 0.4520 \\

Ours-w/o-vae
& \textcolor{red}{0.6116} & \textcolor{red}{64.03} & \textcolor{red}{0.4736}
& \textcolor{red}{0.5728} & \textcolor{blue}{55.60} & \textcolor{red}{0.4023}
& \textcolor{blue}{0.6609} & \textcolor{blue}{62.14} & \textcolor{blue}{0.4612} \\
\rowcolor{gray!15}
Ours
& \textcolor{blue}{0.5893} & \textcolor{blue}{62.96} & \textcolor{blue}{0.4637}
& \textcolor{blue}{0.5630} & \textcolor{red}{55.79} & \textcolor{blue}{0.3998}
& \textcolor{red}{0.6932} & \textcolor{red}{63.32} & \textcolor{red}{0.4945} \\
\bottomrule
\end{tabular}}
\end{table*}

\textbf{Effectiveness of semantic-enhanced visual guidance.}
We compare different image encoder configurations under varying degradation constraints to evaluate the impact of semantic-enhanced visual guidance. As shown in Table~\ref{tab:perception_balance_compare}, the models equipped with semantic guidance (\emph{Ours-w/o-vae} and \emph{Ours}) achieve clear improvements on perceptual metrics such as MUSIQ, MANIQA, and CLIP-IQA, indicating that semantic-enhanced visual guidance enriches perceptual realism and is robust to LR degradation.

\begin{table}
\centering
\caption{Ablation on spatial-grounded textual guidance.}
\label{tab:fidelity_balance_compare}
\resizebox{0.9\columnwidth}{!}{
\begin{tabular}{c|cc|cc|cc}
\toprule
 & \multicolumn{2}{c|}{RealSR} 
 & \multicolumn{2}{c|}{DRealSR}
 & \multicolumn{2}{c}{DIV2K-Val} \\
\cmidrule(lr){2-3} \cmidrule(lr){4-5} \cmidrule(lr){6-7}
Method
& PSNR$\uparrow$ & SSIM$\uparrow$
& PSNR$\uparrow$ & SSIM$\uparrow$
& PSNR$\uparrow$ & SSIM$\uparrow$ \\
\midrule
Ours-w/o-spatial 
& 23.90 & 0.6897
& 26.72 & 0.7406
& 20.70 & 0.5204 \\
\rowcolor{gray!15}
Ours
& \textcolor{red}{25.74} & \textcolor{red}{0.7306}
& \textcolor{red}{28.97} & \textcolor{red}{0.7826}
& \textcolor{red}{21.31} & \textcolor{red}{0.5340} \\
\bottomrule
\end{tabular}}
\end{table}

\textbf{Effectiveness of spatial-grounded textual guidance.}
We next assess the impact of spatial-grounded textual guidance by comparing models with and without spatial guidance under the full semantic enhanced image encoder setting. Table~\ref{tab:fidelity_balance_compare} confirms that adding spatial-grounded textual guidance can improve fidelity metrics (PSNR/SSIM) a lot. As shown in Fig.~\ref{fig:Ablation_vis}, the \emph{Ours-w/o-spatial} variant often introduces hallucinated textures (highlighted in yellow boxes), such as dashed strokes(a), artificial characters appearing on the lantern surface(b), overly detailed and fabricated surface patterns(c), and distorted letter shapes(d), which are inconsistent with the ground truth. In contrast, \emph{Ours-w/o-semantic}, \emph{Ours-w/o-vae}, and \emph{Ours} significantly suppress these artifacts and better align with the ground truth through the spatial-grounded textual guidance.
\section{Conclusions}
\label{sec:concl}
\vspace{-4pt}
We present SpaSemSR, a spatial-semantic guided framework for image super-resolution that explicitly addresses the perception-distortion trade-off. By integrating spatial-grounded textual guidance with semantic-enhanced visual guidance, our approach leverages powerful diffusion priors while mitigating distortion and hallucination. The proposed spatial-semantic attention mechanisms enable adaptive fusion of multimodal priors, producing reconstructions that are both perceptually realistic and structurally faithful. 
Extensive experiments demonstrate that SpaSemSR attains state-of-the-art perception-distortion balance across multiple benchmarks by leveraging spatial and semantic guidance.



\newpage

{
    \small
    \bibliographystyle{ieeenat_fullname}
    \bibliography{main}
}


\end{document}